\pdfoutput=1
\documentclass[11pt]{article}
\usepackage[table, svgnames, dvipsnames]{xcolor}
\usepackage{makecell, cellspace, caption}
\usepackage{booktabs}
\usepackage{multirow}
\usepackage{enumerate}
\usepackage[]{coling}
\usepackage{times}
\usepackage{latexsym}
\usepackage[T1]{fontenc}
\usepackage[utf8]{inputenc}
\usepackage{fontawesome}
\usepackage{subcaption}
\usepackage{microtype}

\usepackage{inconsolata}

\usepackage{graphicx}
\usepackage{amssymb}
\usepackage{xspace}

\newcommand\blfootnote[1]{%
  \begingroup
  \renewcommand\thefootnote{}\footnote{#1}%
  \addtocounter{footnote}{-1}%
  \endgroup
}

\newcommand*{\ethiollmsmall}{EthioLLM-small\xspace}

\newcommand*{\ethiollmlarge}{EthioLLM-large\xspace}
\newcommand*{\afroxlmrlarge}{AfroXLMR-61L\xspace}
\newcommand*{\afroxlmrlargest}{AfroXLMR-76L\xspace}
\newcommand*{\gemmaseven}{Gemma-1.1-7b-it\xspace}
\newcommand*{\gemmatwo}{Gemma-2b-it\xspace}
\newcommand*{\llamatwo}{LLaMA-2-7b-chat-hf\xspace}
\newcommand*{\llamathree}{LLaMA-3-8B-Instruct\xspace}
\newcommand*{\llamathreeone}{Llama-3.1-8B-Instruct\xspace}

\newcommand*{\ethioemo}{EthioEmo\xspace}

\usepackage{caption}
\usepackage{enumitem}
\title{Evaluating the Capabilities of Large Language Models for Multi-label Emotion Understanding}%EthioEmo MultiEmo

\author{\normalsize Tadesse Destaw Belay$^{1, 2, \ast}$, Israel Abebe Azime$^{3,\ast}$, Abinew Ali Ayele$^{4, 5}$, Grigori Sidorov$^{1}$, \\
\textbf{\normalsize  Dietrich Klakow$^{3}$, Philipp Slusallek$^{3}$, Olga Kolesnikova$^{1}$, Seid Muhie Yimam$^{5}$}
\\
\footnotesize
  $^1$Instituto Politécnico Nacional (IPN), CIC, $^2$Wollo University, $^3$Saarland University, \\
  \footnotesize $^4$Bahir Dar University, $^5$University of Hamburg %\\
}

\begin{document}
\maketitle
\blfootnote{$^\ast$ Equal contribution. Corr. email: tadesseit@gmail.com}
% \blfootnote{tadesseit@gmail.com}
\begin{abstract}
Large Language Models (LLMs) show promising learning and reasoning abilities. Compared to other NLP tasks, multilingual and multi-label emotion evaluation tasks are under-explored in LLMs. In this paper, we present \textbf{EthioEmo}, a multi-label emotion classification dataset for four Ethiopian languages, namely, Amharic (amh), Afan Oromo (orm), Somali (som), and Tigrinya (tir). We perform extensive experiments with an additional English multi-label emotion dataset from SemEval 2018 Task 1. Our evaluation includes encoder-only, encoder-decoder, and decoder-only language models. We compare zero and few-shot approaches of LLMs to fine-tuning smaller language models.
% Our evaluation includes baseline classifications, fine-tuning encoder-only pre-trained language models, and prompting of large language models. 
The results show that accurate multi-label emotion classification is still insufficient even for high-resource languages such as English, and there is a large gap between the performance of high-resource and low-resource languages. The results also show varying performance levels depending on the language and model type.
% \blfootnote{Correspondence to: tadesseit@gmail.com}
\textbf{EthioEmo} is available publicly\footnote{\url{https://github.com/Tadesse-Destaw/EthioEmo}} to further improve the understanding of emotions in language models and how people convey emotions through various languages.
\end{abstract}

\section{Introduction}
% Both emotion classification and sentiment analysis tasks are used to extract feelings expressed in a given text. 
In today's digital age, individuals freely express their feelings, arguments, opinions, and attitudes on websites, micro-blogs, and social media platforms. This situation has increased interest in extracting user sentiments and emotions towards events for various purposes such as decision-making, product analysis, customer feedback analysis, political promotions, marketing research, and social media monitoring \cite{kusal2022review}.

Emotion classification is one of the most challenging NLP tasks, where a given text is assigned to the most appropriate emotion(s) that best reflect(s) the author's mental state \cite{tao2020toward}. It poses more challenges than similar NLP tasks, such as sentiment analysis. The challenges of emotion classification as a task worth exploring include many classes, the possibility of a single text expressing multiple emotions, and the cultural and language differences inherent in interpreting or transferring emotions \cite{kusal2023systematic,wang2024largelanguagemodels}. %These factors make emotion classification a task worth exploring.

%There are two different approaches to classifying emotions \cite{wassa-2023}. Multiclass Emotion Classification (MCEC) \cite{ameer2022multi}, where each instance is associated with only one emotion. Conversely, 
Multi-label Emotion Classification (MLEC) considers all emotions expressed in a text, which is a more challenging but essential NLP task, as a text can express multiple emotions simultaneously \cite{ameer2020multi,deng2020multi}. Multi-label classification enables an instance to have any combination (none, one, some, or all) of labels from a given set of emotions. %, allowing a single emotional expression to be linked to multiple emotions

% In many real-world scenarios, a text can express multiple emotions simultaneously, leading to the need for multi-label learning \cite{liu2023emotion}. Naturally, a text can contain more than one emotion, as people often express their ideas in various ways. For instance, someone may begin by appreciating the positive aspects of a service before commenting on its drawbacks on commercial websites. Users may also address multiple aspects of a product/service. By assigning various emotions to such inputs, we can gain a more comprehensive understanding of customer feedback and take more targeted actions to address their concerns. Therefore, we found that multi-label classification is the best choice for our work due to the restrictiveness of the multiclass approach.

This work intends to create and evaluate a multi-label text emotion dataset for the following Ethiopian languages: Amharic (amh), Afan Oromo (orm), Somali (som), and Tigrinya (tir), with an available English dataset for evaluation. As it makes the task more intricate and reflects the complexity often found in real-world data \cite{liu2023emotion}, we follow the multi-label emotion classification approach. 

The main contributions are summarized as:
\begin{enumerate}[noitemsep]
    \item We introduce EthioEmo, a new multi-label emotion benchmark dataset for four Ethiopian languages.
    \item We explore popular Afri-centric encoder-only models that include most of our target languages in the pre-training phase and show fine-tuning performance. % gains from these models.
    \item We evaluate the effectiveness of popular encoder-decoder and decoder-only models for multi-label emotion classification and examine the role of few-shots in improving the task.
    \item We present detailed results and error analyses across languages, data sources, LLMs, and the effects of the translation test set. %that found the difficulty and cultural importance of the emotion classification task.
\end{enumerate}

% The rest of the paper is structured as follows: In Section \ref{related}, we explore previous works on emotion classification. In Section \ref{sec:dataset}, we detail the process of creating the EthioEmo dataset. Section \ref{sec:experiment} discusses experiments with encoder-only,  decoder-only, and in-context learning models for emotion classification tasks. In Section \ref{sec:results}, we continue discussing the experimental results, leading to a dedicated error analysis in Section \ref{sec:error_analysis}. 

\section{Related Work} \label{related}

Emotion recognition involves identifying an underlying emotional state of individuals based on their verbal and nonverbal cues, including text, facial expressions, body language, and speech \cite{DADEBAYEV20224385,AV2024102218}. LLMs are showing promising results for the downstream NLP tasks. Based on the training setup, the model architecture, and the use cases, LLMs can be broadly classified into encoder-only, encoder-decoder, and decoder-only types. In the past few years, there has been a significant increase in the release of decoder-only LLMs at an industry scale, and extensively used for sentiment analysis \cite{zhong2023chatgpt,sentiment-2024}. %  — used for benchmarks where no downstream application is involved \cite{zhong2023chatgpt}. These decoder-only LLMs  the classification setting for sentiment analysis \cite{sentiment-2024}.
We explore emotion classification works in the following categories. 
~\\\textbf{LLMs for text emotion classification}: \citet{emobench-2024} proposed EmoBench to evaluate the emotional cause recognition of LLMs in English and Chinese. \citet{liu2024emollms} proposed EmoLLMs by fine-tuning various open-sourced LLMs for affect analysis and emotion prediction. However, these works are limited to predicting a single emotion class and lack content from languages other than English. \citet{CageggiRU23} fine-tune MT5 and evaluate FLAN and ChatGPT using few-shot prompting approaches for multi-label emotion prediction. Apart from this, the performance of other LLMs has not been assessed for multi-label emotion prediction.%, particularly for low-resource languages. and data domains, only Twitter(X)
%~\\\textbf{Multiclass emotion classification (MCEC)}: Extensive efforts have been made in the literature to construct MCEC text-based emotion benchmark dataset, a text that has only single label emotion, such as SemEval-2007 \cite{strapparava2007semeval}, WASSA-2017\cite{mohammad2017wassa}, EmoBank \cite{emobank-2017}, SemEval-2019 Task 3 \cite{chatterjee-etal-2019-semeval}, AraEmoCorpus \cite{info12020086}, to name a few. 
~\\\textbf{Multi-label emotion classification (MLEC)}: To predict all possible emotions from a text, the following popular datasets were compiled: GoEmotions \citet{demszky2020goemotions}, Balanced Multi-Label Emotional Tweets (BMET) \citet{huang2019seq2emo}, Romanian emotion dataset (REDv2) \citet{ciobotaru-etal-2022-red2}, Multilingual Emotion Prediction (XLM-EMO) \citet{bianchi2022xlm}, WASSA2023 Shared-Task 2 \citet{wassa-2023}, and SemEval-2024 Task 3 \cite{semeval-2024}. 
Nowadays, MLEC tasks also include the corresponding intensity of each identified emotion, such as SemEval-2018 task 1 \cite{mohammad2018semeval}, multimodal multi-label emotion, intensity, and sentiment dialogue dataset (MEISD) \cite{firdaus-etal-2020-meisd}, and EmoInHindi \cite{singh-etal-2022-emoinhindi}.
~\\\textbf{Emotion for Ethiopian languages}: Emotion detection in the context of Africa generally, for Ethiopian languages specifically, has not been studied yet, except for a few sentiment analysis (negative, positive, neutral) tasks \cite{yimam-etal-2020-exploring,tela2020transferring,muhammad-etal-2023-afrisenti}. 
~\\\textbf{Limitations of existing emotion works}: Although there have been several efforts in constructing benchmark datasets and evaluations for text emotion, the existing efforts have the following shortcomings.

\begin{itemize}[noitemsep]
\item  The emotion research is mainly focused on English or a few other high-resource languages \cite{singh-etal-2022-emoinhindi}. 
\item Textual datasets are mostly taken from a single source, such as either news headline \cite{strapparava2007semeval}, YouTube comments \cite{thai2015}, Twitter (X) tweets \cite{mohammad2018semeval}, SMS \cite{wassa-2023}, or Facebook comments \cite{multi-2024}. We might not get all basic emotions from a single data source, and it is hard to generalize about emotion in this case. 
\item The evaluation experiments are focused on classical machine learning and deep learning approaches \cite{survey-2024}, the current state-of-the-art LLMs' multi-label and multilingual emotional understandings are under-explored.
\end{itemize}
Towards this end, we create a multi-label \ethioemo dataset, which is constructed from various sources (news headlines, Twitter (X) posts, YouTube comments, and Facebook post comments), and each instance is annotated with one or more emotion classes. % based on Ekman's \cite{ekman1992argument} six basic emotion classes plus neutral. 
We also conduct rigorous evaluation experiments that classify emotions in multi-label settings using state-of-the-art encoder-only, encoder-decoder, and open-sourced decoder-only LLMs.

\section{EthioEmo Dataset Construction}
\label{sec:dataset}
This section describes the construction of the \ethioemo dataset in detail. The driving force behind creating this dataset is the lack of an available emotion dataset in Ethiopian/African languages. Evaluating LLMs in multi-lingual and multi-label emotion understanding is another under-explored area. % emotion dataset that describes Ethiopian emotions. 
Moreover, emotion is language, culture, and other circumstances dependent \cite{sailunaz2018emotion}. \ethioemo is a new multi-label emotion dataset for four Ethiopian languages, two languages written in Ethiopic Ge'ez (\texttt{gez}) script (\texttt{amh} and \texttt{tir}) and two languages in Latin script (\texttt{orm} and \texttt{som}). We used Ekman’s \cite{ekman1992argument} six basic emotion labels (anger, disgust, fear, joy, sadness, and surprise) plus neutral class.

 \subsection{Lexicon Collections}
 Lexicon entries are emotion keywords that are used to filter instances from millions of collected corpus for annotation. Based on our previous lexicon creation experiences within the Ethiopian context for sentiment analysis \cite{yimam-etal-2020-exploring} and hate speech \cite{ayele-etal-2023-exploring}, we create a list of lexicon entries for each emotion class and language to ensure that each emotion class dataset is balanced and comprehensive. For example, lexicon entries of Joy emotion are “happy,” “excited,” and “thanks” in English. This is a step to balance the dataset by taking equal proportions from each emotion class for annotation. The  Lexicon entries are adapted from an English source, NRC EmoLex \citet{Mohammad13}, with additional manually created emotion keywords. %We applied a keyword-based instance selection technique to balance the data. %The lexicon were only used to retrieve instances from the source or the collected corpus. 
 We obtain the emotion lexicon entries in the following ways:
 \begin{itemize}[noitemsep]
     \item Translate the English NRC EmoLex \citep{Mohammad13} lexicon into Ethiopian languages with the help of Google Translate and native speaker validations (incorrect translations are discarded). 
     \item Collect additional emotion lexicon entries using nearest neighbors of the emotion lexicon entries from available Word2Vec and FastText word embedding models that include our target Ethiopian languages \citep{introduce-semantics,tadesse-impacts}.
     \item We manually add the remaining basic emotion lexicon for each language and emotion class.
 \end{itemize}
 We used 293 \texttt{amh}, 275 \texttt{orm}, 283 \texttt{som}, and 280 \texttt{tir} emotion query entries for six basic emotion classes. We will open-source these lexicons along with the dataset.

\subsection{Data Collection}

%Social media is the primary source of emotion datasets, such as news portals \cite{strapparava2007semeval}, YouTube \cite{thai2015}, Twitter (X) \cite{mohammad2018semeval}, and Facebook \cite{ASEDS2018,distant2016}. Recent observations indicate a substantial increase in the representation and availability of Ethiopian languages across various social media platforms. Among these, the aforementioned platforms provide a rich source of textual emotions for Ethiopian languages.

The datasets have been collected from various sources such as news portals, X/formerly Twitter, YouTube, and Facebook.

\begin{table}[h!]
    \centering
    \begin{tabular}{lcccc}
    \hline
\textbf{Data sources}&\textbf{amh}&\textbf{orm}&\textbf{som}&\textbf{tir}\\
       \hline
       Twitter (X)  & 2000 & 2700 & 2400 & 3100\\
       Facebook   & 1500 & 600  &900 & 600 \\
       YouTube   & 2000 & 2000 & 2000 & 2000\\
       News headline  & 500 & 500 & 500  & 500 \\
       \hline
       \textbf{Total}  & 6000 & 5800 & 5800  & 6200 \\
       \hline
    \end{tabular}
    \caption{Data sources and sample amount taken from each source: Twitter (X) posts, Facebook post comments, YouTube video comments, and news headlines.}
    \label{tab:source}% We randomly select the given amount of instances from each source.
\end{table}

The sources are selected since they have been common data sources for previous emotion classification works and contain rich content for Ethiopian languages \cite{mohammad2018semeval,multi-2024}. These diverse sources are selected to rate the emotions that persist within texts across the sources. The statistics of the data and the sources are presented in Table \ref{tab:source}. As part of the data preprocessing technique, language detection is applied using GeezSwitch \cite{fgaim2022geezswitch} for Ge'ez scripts and pycld3\footnote{\url{https://pypi.org/project/pycld3/}} for Latin scripts languages. We masked user names and URLs to prevent data privacy and confidentiality. For annotation, we select text length with a minimum of 15 characters and a maximum length of a tweet (280 characters). 

%Table \ref{tab:source} shows data statistics we sampled from the most common sources. These sources were selected because they were the most common data sources for previous emotion-related works and contain rich content for Ethiopian languages. These diverse sources are selected to measure also the amount of emotion in these sources. As a part of the data preprocessing technique, language detection is applied using GeezSwitch \cite{fgaim2022geezswitch} for Ge'ez scripts and pycld3\footnote{https://pypi.org/project/pycld3/} for Latin scripts languages. We anonymized users' names and URLs to prevent data privacy and confidentiality. For annotation, we select text length with a minimum of 15 characters and a maximum length of a tweet (280 characters). 

Regarding the period of the collected data, for Facebook, comments from posts between September and December 2023 were extracted as the data was collected at this time using the comment scraper tool\footnote{\url{https://exportcomments.com/}}. For news headlines, we pulled all available BBC (\url{https://www.bbc.com/x}, where x is the name of the language) news headlines using Python script\footnote{\url{https://github.com/keleog/bbc\_pidgin\_scraper}}. For Twitter (X), we used data scraped from 2014 to 2022 using Twitter API for academic research. % \cite{introduce-semantics}. 
For YouTube, we did not consider time span; we collected comments under the playlist/video with the most comments for the specific language using YouTube API. We applied text preprocessing such as language detection, username, URL anonymization, and over-repeated character normalization.

\subsection{Data Annotation}
%For each language, we prepared annotation guidelines and some example text with emotion labels/classes that were provided for the annotators. %The guideline was given to the annotators in addition to a one-day extensive training and follow-up. 
%The data annotation process is done by hiring native speakers for each language. Annotators were provided annotation guidelines with text examples and emotion label(s), hands-on practical training, and pilot tests before the main annotation. The details backgrounds of the annotators are shown in Appendix \ref{sec:appb}. %, Figure \ref{fig:annot}.% Figure \ref{fig:annot} shows details of the annotators' backgrounds. %In total, 23 native speaker experts participated in the annotation. 

For the data annotation, we employed native speakers for each language. Annotators were provided annotation guidelines with text examples and emotion label(s), hands-on practical training, and pilot tests before the main annotation. We compensated annotators with a payment of roughly \$6 per hour on average, nearly the same as the hourly wage of Master's degree holders in Ethiopia. The detailed backgrounds of the annotators are shown in Appendix \ref{sec:appb}.

We customize the POrtable Text Annotation TOol (POTATO) \cite{pei-etal-2022-potato} for our in-house annotation platform. A minimum of three annotators annotated each instance.  %Appendix \ref{appendix:annotation_tool} shows a screenshot of the annotation tool.
The data is annotated in multiple batches by assessing the data quality and annotators' performance, including control questions and agreements in each batch.  Disagreed instances were re-annotated by new annotators, and if no agreement was reached again, the instances were excluded from the dataset. The final gold label was determined based on agreement by at least two annotators for each emotion class. %Annotators received a moderate remuneration for their work. %For the final \ethioemo dataset, 22,752 instances were annotated by three or more annotators.

\subsection{Inter-Annotator Agreement (IAA)}
The most common IAA measurements, such as Cohen’s kappa \cite{cohen1960coefficient}, Fleiss’ kappa \cite{fleiss1971measuring}, Krippendorff’s alpha \cite{krippendorff2011computing}, and bootstrapping method \cite{marchal2022establishing} do not support multi-label with multiple annotators at the same time.  We adopted a multi-label agreement (MLA) method proposed by \citet{li2023annotation} to obtain the multi-label agreement among all annotators. We also computed free marginal Randolph's Kappa scores \cite{Randolph2005FreeMarginalMK}, a metric well-suited for measuring inter-annotator agreement in tasks involving multiple annotators.

\begin{table}[h!]
    \centering
    \begin{tabular}{lccc}
        \hline
\textbf{Language}  & \textbf{MLA} & \textbf{Cohen's K.} & \textbf{Free M.}\\
       \hline
        Amharic    & 0.50  &0.52 & 0.65\\
        Afan Oromo & 0.64  &0.66 & 0.76\\
        Somali     & 0.51  &0.50 & 0.66\\
        Tigrinya   & 0.53  &0.57 & 0.68\\
        \hline
    \end{tabular}
    \caption{IAA of the \ethioemo Dataset. Multi-label Agreement (MLA) is a direct agreement between the multi-label classes and all annotators. Cohen's Kappa is a pairwise agreement between two annotators, and the result is an average pair-wise of the three annotators. Free Margin (Free M.) is calculated as a pairwise agreement between two emotion classes and takes an average.} 
    \label{tab:iaa}
\end{table}

According to the work of \citet{sanchez2016let}, the IAA results in Table \ref{tab:iaa} show moderate and above agreement as Cohen's Kappa score ranges from 0.41-0.60 is moderate. For further analysis of IAA, we observe Cohen's kappa agreement for four main emotion classes (Anger, Disgust, Sadness, and Joy), and the results are Amharic: 0.74, Afan Oromo: 0.81, Somali: 0.75, and Tigrinya: 0.77, showing significantly higher scores than the scores obtained from the total of seven classes. This shows that IAA scores vary with the number of classes, as more classes generally increase the complexity of annotation, often lowering agreement scores \cite{piskorski-2023-holistic}. Based on Cohen's Kappa value, our IAA result is also comparable with related works, as the GoEmotion \cite{demszky2020goemotions} dataset IAA is 0.29 for 27 emotion classes. %While this data has 27 classes, we have a high comparable agreement result. 
The highest agreement scores are reported for \texttt{Afan Oromo}. We manually go through the annotator-level data and observe that most of the annotators selected single emotions during the annotation, the reason for \texttt{Afan Oromo} having a better agreement score. This shows that the number of annotated labels by each annotator is inversely proportional to the agreement score. The overall IAA agreement score shows multi-label emotion task difficulty, a condition where an instance can have none, one, two, or all emotion classes with multiple annotators.

\section{Evaluation Settings}
\label{sec:experiment}
Training and testing LLMs such as GPT-4 \cite{openai2024gpt4}, Mixtral 8x22B \cite{jiang2024mixtralexperts}, PaLM-340B \cite{palm2technicalreport}, and LLaMA-405B \cite{llama3.1} are often not feasible for academic researchers and companies with limited resources. As a result, there has been a shift towards smaller language models \citep{smallmodelsllm-2024}. Our experiment includes pre-trained encoder-only, encoder-decoder, and medium-size parameter decoder-only models for scientific reproducibility. We fine-tune encoder-only models using the \ethioemo training dataset and evaluate zero-shot and in-context learning predictions with LLMs.
% We perform several experiments with different approaches to provide a robust baseline experiment for the \ethioemo dataset. 
% In our evaluation, we include the English multi-label emotion dataset from SemEval 2018 Task 1: Affect in Tweets dataset \cite{mohammad2018semeval}. %We excluded the four least proportioned emotion classes (anticipation, love, optimism, and pessimism) from the English dataset to align with our dataset, which is six basic emotions.  
The statistical distribution of the \ethioemo and English datasets is shown in Table \ref{tab:train_test}. %In our experiment, we exclude \texttt{Neutral} class; all the results are with the six basic emotion classes.

\begin{table}[h!]
    \centering
%Neutral class is excluded in the train-test split %We randomly stratified and split our dataset into train (60\%), dev (10\%), and test (30\%) sets.}% The differences from data source Table \ref{tab:source} are disagreed instances (instances that have no label agreement between annotators) are excluded}
    \begin{tabular}{lcccc}
    \hline
\textbf{Language}&\textbf{Train}&\textbf{Test}&\textbf{Dev}&\textbf{Total}\\
       \hline
       % Spanish  & 3561 & 2854  &679 & 7,094 \\
       % Arabic  & 2278 & 1518 & 575 & 4,371\\
       Amharic     & 2,614 & 1,309 & 437 &4,360 \\
       Afan Oromo  & 2,598 & 1,300 & 435 &4,333 \\
       Somali      & 2,087 & 1,045 & 349 &3,481\\
       Tigrinya    & 2,865 & 1,435 & 479 &4,779\\
       English     & 6,327 & 1,232 & 845 & 8,404\\
       \hline
    \end{tabular}
    \caption{Statistics of train, test, and dev sets for \ethioemo along with SemEval-2018 Task 1 English dataset. We randomly stratify to split the \ethioemo dataset into train (60\%), dev (10\%), and test (30\%) sets. These statistics are without the \texttt{Neutral}  class as our overall experiments do not include \texttt{Neutral} class in the evaluation. Final annotated dataset statistics with \texttt{no emotion} or \texttt{Neutral} class are Amharic: 5,891, Afan Oromo: 5,690, Somali: 5,631, and Tigrinya: 6,109, a total of 23,321 instances were annotated.}
    \label{tab:train_test}
\end{table}

% Our experiment includes both encoder-only and decoder-only architecture models. We fine-tune various encoder-only models using the \ethioemo training dataset. Furthermore, we conduct evaluations using zero-shot and few-shot predictions with LLMs. 
%Do not describe sequences in academic writing. ... we first do this, then this, then this... finaly this ... This is bad writing style. It looks like execution of a pre-determined process. Rather, focus on the main tasks and the reader should deduce the sequence bythemselves 

\subsection{Afri-centric Encoder-only Models}%Experiments
\label{sec:encoder-only}
Considerable efforts have been dedicated to creating multilingual BERT-based encoder-only models for African languages. We select encoder-only models based on popularity, and models include at least two languages from our target languages. %Our experiments include zero-shot and fine-tuning the models using the train-test split, shown in Table \ref{tab:train_test}.%, shown in Table \ref{tab:baseline}.%, as shown in Figure \ref{fig:zero-shot-all},

We make zero-shot and fine-tuning evaluations using the following Afri-centric pre-trained language models.
\textbf{AfriBERTa} \cite{ogueji-etal-2021-small} pre-trained on 11 African languages. It includes our four target languages.
\textbf{AfroLM} \cite{dossou2022afrolm}: a multilingual model pre-trained on 23 African languages, including \texttt{amh} and \texttt{orm} from Ethiopian languages. 
\textbf{AfroXLMR} \cite{adelani-etal-2024-sib}: adaptation of XLM-R-large model \cite{conneau2019unsupervised} (has two versions: 61 and 76 languages) for African languages including the four Ethiopian languages and high-resource languages (English, French, Chinese, and Arabic). 
\textbf{EthioLLM} \cite{tonja2024ethiollm}: multilingual models for five Ethiopian languages (\texttt{amh}, \texttt{gez}, \texttt{orm}, \texttt{som}, and \texttt{tir}) and English.

\subsection{Open Source Decoder-only Models}
\label{sec:zeroshot}
From the family of decoder-only LLMs, we work with instruction-tuned versions of popular open-source models. Namely, Llama-2-7b \cite{touvron2023llama}, Llama-3-8B \cite{metaIntroducingMeta}, Llama-3.1-8B \cite{llama3.1}, Gemma-1.1-7b \cite{gemmateam2024gemma}, and Gemma-2b \cite{gemmateam2024gemma}. From encoder-decoder, we evaluate Aya-101 \cite{aya-2024} — fine-tuned from mT5 \cite{xue-etal-2021-mt5}.  It is a multilingual 13B parameter model that follows instructions in 101 languages, including \texttt{amh} and \texttt{som}. We choose these models based on their popularity in the open-source community and serve as a baseline for similar NLP task evaluation. From closed-source LLMs, we include GPT-4o-mini in our evaluation as it is cost-efficient and easy to reproduce \cite{gpt4o-mini}. We used English-based prompts for evaluating LLMs following the work by \citet{zhang2024impact,agarwal2024many} as English prompts work better than in-language prompts. %Details of the prompts are shown in Appendix \ref{appendex:zerofewshot}.

\subsection{Translate Test Experiments}
\label{sec:translate-test}
Following the work by \citet{etxaniz2023multilingual}, one approach to improve the performance of multilingual language models is to translate the data to English using existing machine translation systems. Our approach involves translating the \ethioemo test dataset to English to determine if English-centric models can solve the task efficiently. For the translation, we used the NLLB-200-3.3B multilingual machine translation model \cite{nllbteam2022language}. %Given that our dataset is collected from different sources that reflect the cultural values of the speakers, this task aims to demonstrate the difficulty of using models that were not trained on these datasets directly.

\subsection{In Context Learning (ICL)}
One approach to improve the performance of LLMs is to show them examples of the task. Following the work of \citet{zhang2024impact,agarwal2024many}, we use in-context learning to teach the models about the task without parameter updates by showing them input and output examples. We work with 2, 4, 6, and 8 demonstrations in our experiment and compare them with zero-shot experiments. We increase the number of contexts (k shots) by two to show the slightly increasing effects of examples. For our k-shot experiments, we applied randomly selected in-language examples from the dev set, which remained consistent across models.
%Given the size of open-source decoder-only models, it is hard to fine-tune each model for a specific task.
% \subsection{Evaluation Metrics and Methods} 
% We used automatic evaluation metrics such as weighted-averaged F1-score \cite{manning_introduction_2008}. We report this metric as it is calculated by taking the mean of all class F1-scores while considering each class's support, and it is more convenient for multi-label classification and class-imbalanced data. Other multi-label evaluation metrics such as multi-label accuracy, macro F1-score, and micro F1-score are also reported in Appendix \ref{sec:appendix2}.
We used log likelihood-based evaluations using lm-evaluation-harness\footnote{\url{https://github.com/EleutherAI/lm-evaluation-harness}} by \citet{eval-harness} for zero-shot and few-shot LLMs experiments.
% \footnote{https://github.com/EleutherAI/lm-evaluation-harness}

\section{Results}%Experimental
\label{sec:results}
\subsection{Fine-tuned Encoder-only Models}
 Results of fine-tuned encoder-only models are shown in Table \ref{tab:baseline}. Based on the results, \afroxlmrlargest outperforms for \texttt{amh} and \texttt{orm} with a 69.9\% and 72.6\% F1 score, respectively, as both languages are included in the pre-training. Examining the overall encoder-only results, AfroXLMR families perform better for the target languages. We observe that languages included in the pertaining phase perform better. Although encoder-only models demand more training data and computational resources, they still have significant room for improvement in tackling multi-label emotion classification tasks. \textbf{Pre-training is important for multi-label emotion classification task}. This is evidenced by the F1-scores we present, where the highest score is achieved by the \texttt{orm} language from \afroxlmrlargest with a score of 72.6\%. 

\begin{table}[h!]
    \centering
    % \resizebox{\columnwidth}{!}{%
    \begin{tabular}{lcccc}
    \toprule
        \textbf{Model name} & \textbf{amh} & \textbf{orm} & \textbf{som} & \textbf{tir}\\% & average\\
        \hline
        \multicolumn{5}{l}{\textit{Fine-tuned encoder-only models}} \\
         \ethiollmsmall  &65.3 &69.4  &47.1  &55.7\\% &56.3\\
         % EthioLLM-l-70K  &58.7 &64.6  &44.2  &56.4 \\
         \ethiollmlarge &64.2 &67.4  &38.0 &56.7\\% &50.6\\
         \afroxlmrlarge &68.3 &66.5  &\textbf{64.2}  &\textbf{62.4}\\% & 61.0\\ 
         \afroxlmrlargest &\textbf{69.9} &\textbf{72.6}  &62.6 &58.1\\% & \textbf{61.6}\\
         AfroLM-active-l &65.4 &67.7  &\cellcolor{Gainsboro!80}52.0  &\cellcolor{Gainsboro!80}53.2\\% &47.9\\
         AfriBERTa-large &51.6&71.4  &63.2  &60.7\\% &61.1\\
         \bottomrule
    \end{tabular}
    % }
    \caption{Weighted-averaged F1-score results from fine-tuned pre-trained language models. The \colorbox{Gainsboro!80}{light-gray} shows the model does not include the languages in the pre-training.}
    \label{tab:baseline}
\end{table}
%Additionally, the best average score across languages is achieved by \afroxlmrlargest, with 61.6\% F1-score.

\subsection{Zero-shot Experiments} 
% \textbf{Prompts}: , We perform zero-shot evaluations and keep all prompt templates in English. For encoder-only, decoder-only, as previous studies show better performance with English prompts on LLMs than in-language prompt \cite{lin-etal-2022-shot,muennighoff-etal-2023,shen-etal-2024}. 
 % and translate test experiments that are discussed in Section \ref{sec:encoder-only}, \ref{sec:zeroshot}, and Section \ref{sec:translate-test} respectively, 
We conduct a zero-shot evaluation and make the following observations, as summarized in Table \ref{tab:zero-shot-all}.

\begin{table*}[h!]
\centering
% and LLaMa families are primarily English models. We used the full HuggingFace name of the models for ease of understanding.
% \resizebox{\columnwidth}{!}{%
\begin{tabular}{llllllc}
\toprule
 \textbf{Pre-trained LMs} &{\textbf{amh}} & {\textbf{orm}}  &{\textbf{som}} & {\textbf{tir}}& {\textbf{eng}} & \textbf{Average} \\
\midrule
\multicolumn{6}{l}{\textit{Zero shot for encoder-only}} \\ 
\ethiollmsmall & 31.72 & 12.88 & 30.88 & 32.09 & 37.76 & 29.07 \\
% EthioLLM-l-70K & 24.81 & 16.56 & 9.53 & 19.71 & 27.90 \\
\ethiollmlarge & 14.38 & 32.35 & 10.53 & 10.94 & 37.87 & 21.21\\
\afroxlmrlarge &22.05 & 39.68 & 21.12 & 22.30 & 43.00  &  29.63\\
\afroxlmrlargest & 28.62 & 35.81 & 14.86 & 15.94 & 24.38 & 23.92\\
AfroLM-active-l & 39.91 & 25.60 & \cellcolor{Gainsboro!80}15.92 & \cellcolor{Gainsboro!80}\textbf{35.63} & 33.93 & 30.20\\
AfriBERTa-large & 25.67 & 15.57 & 19.98 & 35.39 & 26.38 & 24.60\\
\midrule
\multicolumn{6}{l}{\textit{Zero shot for decoder-only}} \\ 
\gemmatwo &10.22 & 7.81 & 14.26 & 7.87 &  46.4 & 17.37\\
\gemmaseven &27.94 & 34.87 & 25.87 & 19.55 &  65.73 & 34.79\\
\llamatwo & 17.35& 19.24 & 22.05& 12.97 &  54.07 & 25.14 \\
\llamathree &28.18 & 26.91 & 29.29 & 19.73 &66.74 & 34.17 \\
\llamathreeone & 20.58 & 24.10 & 22.07 & 10.28 & 51.17 & 25.25\\
% \midrule
% \multicolumn{6}{l}{\textit{Zero shot for multilingual decoder-only }} \\ 
Cohere-aya-101 & 48.80 & 33.65& 43.00& 38.97 &66.20  & 46.12\\
%\rowcolor{Gainsboro!80}
\midrule
\multicolumn{6}{l}{\textit{Zero shot for closed models}} \\ 
GPT-4o-mini & 53.86 & 47.84& 52.04 & 35.47 & 70.98 & 52.04\\
\midrule
\multicolumn{6}{l}{\textit{Zero shot for decoder-only translated to English }} \\ 
\gemmatwo & 30.91& 28.66& 31.43  &22.54 & &28.39\\
\gemmaseven & 45.05 & 47.86 & 44.72  & 35.35 & & 43.25\\
\llamatwo & 35.48& 34.27 & 34.58  & 25.19 & & 32.38\\
\llamathree & 48.26 & 48.55 & 46.46 & 40.93 & & 46.06\\
\llamathreeone & 28.59& 34.43 & 32.28 & 21.66 &  & 29.24\\
% \midrule
% \multicolumn{6}{l}{\textit{Zero shot for Multilingual decoder-only translated to English }} \\ 
Cohere-aya-101 & 44.95 & 43.39& 41.52 & 31.63 &  & 40.37\\
\midrule
\multicolumn{6}{l}{\textit{Translated zero shot for closed models}} \\ 
GPT-4o-mini & 55.89 & 51.59& 51.14 & 47.60&  & 51.56\\

\bottomrule
\end{tabular}
% }
\caption{Zero-shot experiment results from encoder-only and decoder-only models (weighted-averaged F1-score) across languages. The translated test is by translating \ethioemo test set to English. The \colorbox{Gainsboro!80}{light-gray} background indicates AfroLM does not include the languages in the pre-training. }
\label{tab:zero-shot-all}
\end{table*}

\textbf{Encoder-only models still have an advantage over the recently popular open-source decoder-only models for low-resource languages}. We compare zero-shot results of LLMs with zero-shot and fine-tuned encoder-only models, and LLMs under-perform compared to encoder-only models. This is likely due to their initial multilingual setup of encoder-only models for low-resource languages. Cohere-aya-101 outperforms all decoder-only models with an average score of 46.12\% as it is designed for multilingual and officially includes \texttt{amh} and \texttt{som} languages. Looking at the target languages, the closest performance we see between encoder-only and encoder-decoder models is in the \texttt{amh} language, with a score difference of 8.9\% between AfroLM and Cohere-aya-101. For the encoder-only model, we can see \afroxlmrlargest takes the lead, which explains its top performance in the fine-tuning experiment. From zero-shot evaluations, Cohere-aya-101 consistently outperforms in all languages except \texttt{orm}. In general, the result shows how fine-tuning smaller and more efficient pre-trained language models can still outperform zero-shot performances of LLMs, which have room for improvement in multi-label emotion classification. Considerably, zero-shot or in-context learning of LLMs is not comparable with the BERT family’s pre-trained model that has already seen the language in the pre-training phase. However, we compare only according to the resources that LLMs consume to fine-tune, and we expected LLMs to perform better based on their size.% The primary reason \gemmaseven avoids this under-performance is its high scores in English. for \texttt{orm} AfroXLMR-61L is better

\subsection{Translate Test Experiments}
\textbf{The models struggle to classify multi-label emotions even after translating the test set to English}.  We conduct an experiment using the translation of the test set to investigate the reasons for poor performance in decoder-only models, as discussed in Section \ref{sec:translate-test}. Our findings reveal that even after the test set is translated into English, these models still struggle to identify emotions accurately compared to English. 
%, and different language scripts will not be a problem.   possibly due to its larger vocabulary size and facilitating better vocabulary sharing across multiple languages.
In particular, Cohere-aya-101 performs poorly in all \ethioemo translation test set evaluations compared to a near similar size LLaMA-3-8B-Instruct model. This might be due to either the limitations of the machine translation system employed (we do not have ground truth for further translation quality checking) or the inherent complexities of the emotion task that may not carry the same meaning across languages in the translation.  
% However, Cohere-aya-101 demonstrated better results in the translation experiment compared to the rest, as it is a multilingual fine-tuned encoder-decoder model.

\subsection{In-Context Learning Results}
We do in-context learning experiments because fine-tuning LLMs can incur enormous computing costs. This approach helps improve the model's understanding ability without any parameter update. 

\begin{figure*}[!ht]
    \centering
    \includegraphics[width=\linewidth]{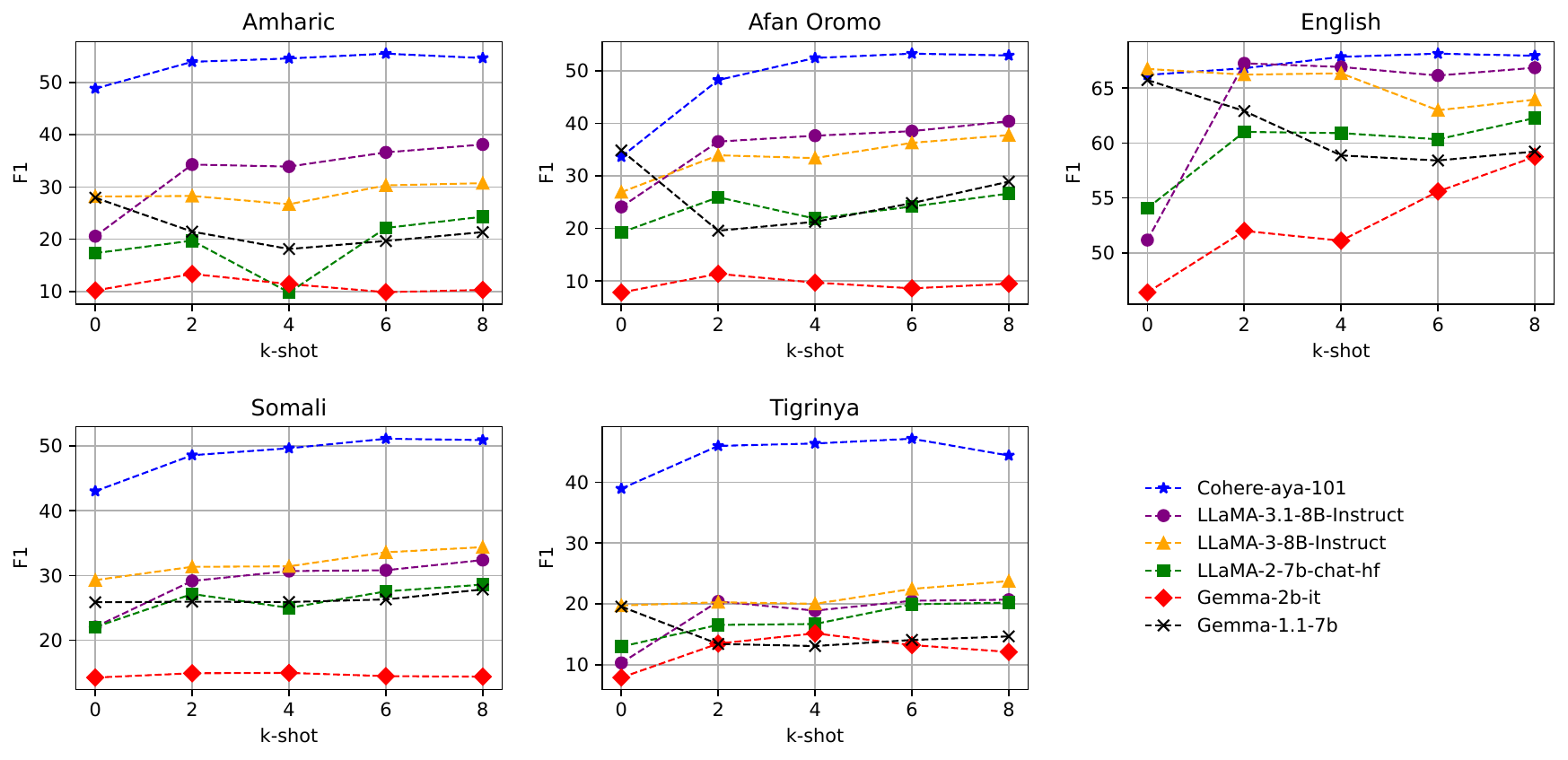}
    \caption{In-context learning (ICL) experiments with k-shots and languages.}
    \label{fig:zero-shot-all}
\end{figure*}

\textbf{All models benefit from two-shot examples compared to zero-shot tests}. Based on the results shown in Figure \ref{fig:zero-shot-all}, all our models benefit from two-shot contexts.  Looking at the Ethiopian languages, we can see that they all improved their scores by showing two examples compared to zero-shot tests. However, this improvement is not shown in \gemmaseven, which already had good scores in the zero-shot experiment across languages. %This might be due to the Latin script that Afan Oromo uses, providing a better cross-lingual understanding evidenced by zero-shot scores as shown from Table \ref{tab:zero-shot-all}.  
Among target languages, \texttt{orm} gains the highest scores in the zero-shot experiment with \gemmaseven model.  The same pattern does not apply to \texttt{som} — it uses Latin script as \texttt{orm}, which requires further investigation. For English, \gemmaseven at four-shots has a better comparable result to the zero-shot.% four shot achieves better result with 58.9\% F1 score.% baseline result \cite{mohammad2018semeval}. %This issue will be discussed in the error analysis section, Section \ref{sec:error_analysis}. %\israel{remove this sentence if we dont have explanation}58.9

\textbf{Examining the impact of increasing the number of shots by two examples is not guaranteed to improve performance}. We observed that the improvement was inconsistent and could not be guaranteed. However, there are clear performance gains from 0 to 2 shots, 2 to 6, 2 to 8, and 4 to 8  shots. This is particularly evident in all languages. The encoder-decoder Cohere-aya-101 model has a comparable best result for low-resource languages to the commercial GPT-4o-mini. \textbf{Encoder-decoder Cohere-aya-101 model outperforms the open-source LLMs}.   The exception for the lowest performance for \texttt{tir} is that it is not included in the pre-training of the Cohere-aya-101 model.  Another observation is that improvements in results are associated with the sizes of the models' size or parameters, such as from Gemma-2b-it to Gemma-1.1-7b and from LLaMA-2-7B to LLaMA-3-8B. %even for the languages not officially included in the pre-training, such as \textit{orm}. commercial GPT-4o-mini

% \textbf{Data sources from news headlines are mostly \textit{Neutral}, while other sources are better for the basic emotions.} We analyze the data sources as shown in Appendix \ref{sec:app-datasource}; instances sourced from news headlines are almost \textit{Neutral}. Emotion classes such as \textit{Anger}, \textit{Disgust}, \textit{Joy}, and \textit{Neutral} are shared on Facebook comments and Twitter (X) posts. YouTube comments contain nearly all basic emotions; a better source for Ethiopian language emotion data, including the rare \textit{Surprise} emotion class. % Twitter (X) posts, Facebook comments, and YouTube comments are good sources for the basic emotions}. 

\subsection{Prompt Sensitivity Experiment}
\textbf{Role-based prompt gained more results from LLMs}. A drawback of the prompting evaluation is the model sensitivity to prompts, where slight changes in instruction can lead to large differences in performance \citep{sun2023evaluating}. To handle this, we use the following three prompts: (1) \textbf{generic}: a prompt which does not give information about the task, used in \cite{liu2024emollms}; (2) \textbf{task-based}: describes the given task \cite{edwards-2024}; (3) \textbf{role-based}: a new prompt which gives more information, including \texttt{"You are a helpful AI assistant that can identify emotions from text"}. All prompting results presented in this paper are averages of the three prompts. For reproducibility of the experiment, the prompts are shown in Appendix \ref{fig:prompt}, and the results of each prompt are presented in Appendix \ref{sec:prompt}.

\begin{figure}[hbt!]
    \centering
    \includegraphics[width=\linewidth]{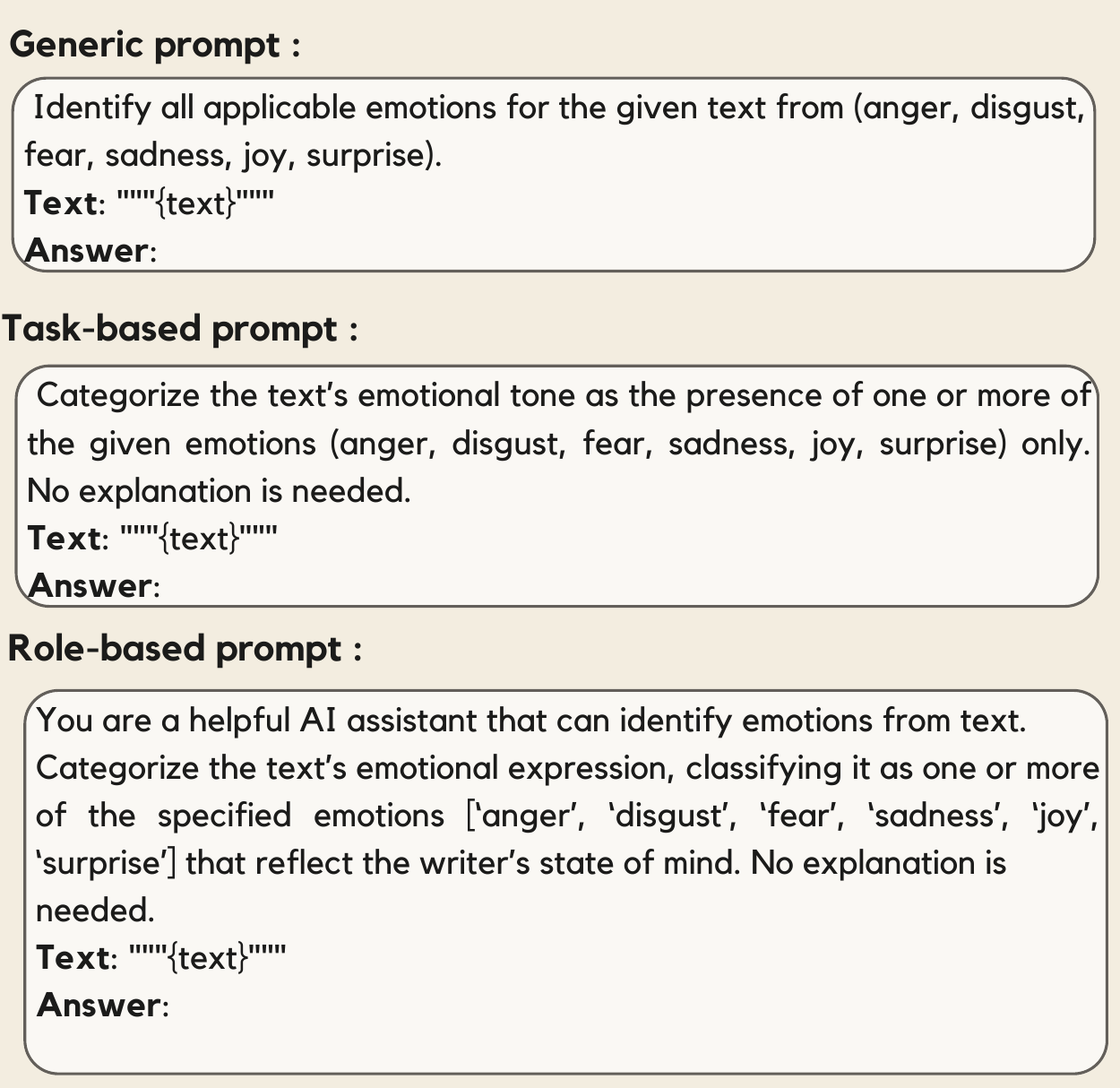}
    \caption{The three prompts used for decoder-only zero-shot and in-context learning experiments}
    \label{fig:prompt}
\end{figure}

\section{Error Analysis and Discussion}
\label{sec:error_analysis}

\paragraph{Task difficulty:} Our analysis shows that the task is not easily solvable by any of the methods. This shows the significance of this task in evaluating the existing models and observing that multi-label emotion classification needs more exploration, even for high-resource languages such as English. Some of the difficulties include 1) inability to know the exact feelings of the writers in sarcastic texts — needs context, and 2) ambiguity between some emotion classes such as \texttt{Anger} and \texttt{Disgust} (for example, in 61 instance annotations, \texttt{Anger} and \texttt{Disgust} appear together from 85 disagreed \texttt{amh} instances). %, and 3) differences in human experience that impact how they perceive emotion in text. 
The fact that the task is difficult also means a great deal of work to advance further research in emotion detection and analysis, which shows that \ethioemo dataset is a useful contribution to evaluating the upcoming advanced models.

\textbf{Data sources from news headlines mostly exhibit none of the six basic emotion classes, while other sources are better for the basic emotions.} We visualize the statistics of emotion distributions across languages and data sources are shown in Appendix \ref{sec:app-datasource}; instances sourced from news headlines are almost \texttt{Neutral - do not have any of the basic emotions}. Emotion classes such as \texttt{Anger}, \texttt{Disgust}, and \texttt{Joy} are shared on Facebook comments and Twitter (X) posts. YouTube comments include all basic emotions — a better source for Ethiopian languages' emotion data that has the rare \texttt{Surprise} emotion class. The statistics of emotion distributions across languages and data sources are shown in Appendix \ref{sec:app-datasource}. % Twitter (X) posts, Facebook comments, and YouTube comments are good sources for the basic emotions}. 
\paragraph{Challenges in emotion annotation :} Obtaining consistent annotations for an NLP dataset, especially for emotion, is challenging. This is due to several reasons, including 1) difficulty in knowing the exact feeling of the writers in sarcastic texts, 2) differences in human experience that impact how they perceive emotion in text, 3) sometimes annotators depend only on emotion keywords present in the text during annotation, 4) ambiguity between some emotion classes such as \textit{Anger} and \textit{Disgust}, and 5) reports of third person sayings as the writer's emotion are some of the challenges encountered during the annotation.

\paragraph{Experiment error analysis:} 
Regarding the dataset emotion distribution, the dataset has more \texttt{Anger} and \texttt{Disgust}. Firstly, this is common also in other emotion datasets \cite{mohammad2018semeval,demszky2020goemotions,semeval-2024}. Secondly, one reason is because of the conflict situations (in the year 2023) in some parts of Ethiopia and in the global context, such as the Hamas-Israel and Russia-Ukraine wars.

We go through the predicted test file from the best-performing encoder-only model, \afroxlmrlargest,  with the help of experts from each language. %and we identify available cases for the incorrectly predicted instances. %The details of the error analysis with examples are shown in Appendix \ref{sec:app-error}. 
The following are the most common cases of the incorrect prediction of emotions. 1) While the gold labels of an instance have more than one emotion, the models predict a single emotion label and vice versa — a common issue in a multi-label classification problem. 2) The model classifies based on some emotion keyword/emojis in the text, not the whole context. \textbf{This shows that text with emojis is straightforward for the models to predict emotions} and is aligned with the work of \citet{liegl2024emotional}. 3) The model fails due to incomplete text and grammatical errors. This is mainly due to the limited length of tweets and the informal writing style of social media \cite{qa-etal-2022}.  
% 4) Understanding the feelings behind specific texts can be challenging, especially regarding sarcastic remarks or quotes from third person.
%6) While the dataset has no neutral category with other emotions, the model sometimes combines the neutral class with other emotion categories during prediction. 
4) The model fails to categorize a text into specific emotion classes, resulting in nothing predicted. This problem is shown in encoder-decoder models. In multi-label classification, encoder-only works in OneVsRest (One vs other) approach \citep{app12115396} which is predicting whether
each emotion is present or not separately, for instance, \texttt{Anger} or \texttt{Not anger}, \texttt{Fear} or \texttt{Not Fear}. When the model responds \texttt{Not} for all emotions, the result will be nothing predicted. On the other hand, decoder-only models such as GPT-4o-mini \cite{openai2024gpt4} show an over-predicted problem, assigning more emotion classes while most of the instances have a single emotion class. For these above-mentioned cases, emotion instance examples are shown in Appendix \ref{sec:app-error} for the corresponding language and case number. 

\section{Conclusion and Future Work}
In light of the growing interest in creating challenging NLP tasks to assess the abilities of LLMs, language- and culture-specific datasets are becoming crucial \cite{wang2024mmlu, adelani2024irokobench}. This work presented a multi-label emotion dataset (\ethioemo) and an evaluation of multi-label emotional understanding of encoder-only, encoder-decoder, and decoder-only language models. The dataset provides diversity regarding the data source (X/Twitter posts, YouTube comments, Facebook comments, and news headlines) and four Ethiopian languages with available English dataset for evaluation).
 % Given that our dataset is collected from different sources that reflect the cultural values of the speakers aims to demonstrate the difficulty of using models that were not trained on these datasets directly.
We reported strong baseline results using various experimental settings such as fine-tuning encoder-only models, translated test sets, prompt sensitivity, zero-shot, and impacts of increasing the number of shots for in-context learning evaluations. Encoder-only Afri-centric models that include target languages during the pre-training phase are the best for the classifications of the \ethioemo dataset. In general, the results show that fine-tuning encoder-only language models can still outperform the few-shot approaches of LLMs. The open-source Cohere-aya-101 model outperformed other LLMs next to the commercial GPT-4o-mini.
% Before any fine-tuning of LLMs, the first step is to evaluate how the model performs without any fine-tuning to get a baseline for pre-trained model performance.
This paper focused on evaluating state-of-the-art open-source LLMs with the least parameters for scientific reproducibility. Fine-tuning open-source and evaluating closed-source LLMs are out of the scope of this work and are the next works.
We believe this dataset and results can be employed as a baseline in the future for better multi-label emotion classification tasks. Resources such as lexicons, annotation guidelines, and datasets are publicly available for further investigation. 
 %testing the highest parameter versions of LLMs'
%The results also show that there is still a certain gap between the current open-sourced LLMs, GPT-4o-mini in specific domains.

\section*{Limitations}
In this work, we present and evaluate the \ethioemo dataset using Afri-centric pre-trained language models and open-source LLMs. Despite our efforts, the following are limitations of this work.

\noindent\textbf{Imbalanced data}: Even if it is impossible to balance emotion data, we tried to balance the emotions using lexicon entries. One of the limitations of this work is that the distribution of the emotion classes is imbalanced. Having more balanced data would be better. However, the nature of the task itself makes it challenging to balance each emotion class because all emotions are not expressed equally in the data source platforms. %several reasons, including 1) difficulty in knowing the exact feeling of the writers in sarcastic texts, 2) differences in human experience that impact how they perceive emotion in text, 3) sometimes annotators depend only on emotion keywords present in the text during annotation, 4) ambiguity between some emotion classes such as \textit{Anger} and \textit{Disgust} (for example, in 61 instances, \textit{Anger} and \textit{Disgust} appear together with the \textit{Neutral} class from 85 disagreed Amharic instances), and 5) reports of other people sayings as the writer's emotion are some of the challenges encountered during the annotation.

\noindent\textbf{Translation effect on emotions}: To evaluate generative models, we translate the \ethioemo test dataset to English to know if the prediction difficulties come from the task's nature or language understanding. However, this translation will have quality and context effects on the emotion itself as emotions are culture and language-dependent. %, We have observed that the translation of Afan Oromo is poorly done in the NLLB-200-3.3B model. Various prompting styles may also affect the result; we use a single prompt for all evaluated decoder-only models, shown in Appendix \ref{appendex:zerofewshot}.

%Few encoder-only and generative models were evaluated: due to the scale of the experiments, we limited our evaluation to a few pre-trained encoder-only models and decoder-only LLMs.
% \textbf{Data ethical issues:} Some platforms do not allow the data to be publicly available. We use this data only for emotion research by anonymizing private information, such as usernames to \textit{NAME} and URLs to \textit{URL}. %In one or another way, all available emotion datasets are sourced from these platforms.

% Please ensure that Bib\TeX{} records contain DOIs or URLs when possible, and for all the ACL materials that you reference.

\section*{Acknowledgments}

This work was carried out as a part of the AfriHate project of data annotation with the support of the Lacuna Fund, an initiative co-founded by The Rockefeller Foundation, Google.org, and Canada’s International Development Research Center. 
%The work was done with resource support from the Mexican Government through the grant A1-S-47854 of CONAHCYT, Mexico, grants 20241816, 20241819,  and 20240951 of the Secretaría de Investigación y Posgrado of the Instituto Politécnico Nacional, Mexico. 
The authors thank the CONAHCYT for the computing resources brought to them through the Plataforma de Aprendizaje Profundo para Tecnologías del Lenguaje of the Laboratorio de Supercómputo of the INAOE, Mexico, and acknowledge the support of Microsoft through the Microsoft Latin America PhD Award.  We thank OpenAI for providing API credits to Masakhane. We also acknowledge the support of the LT Group, the University of Hamburg, for hosting the annotation tools.

\bibliography{custom}
\onecolumn
\appendix

\section{\ethioemo Languages}
\label{sec:appendix1}
There are more than 2000 languages spoken in the African continent, and more than 80 of them are spoken in Ethiopia\footnote{https://www.statista.com/statistics/1280625/number-of-living-languages-in-africa-by-country/}. %Amharic is the Federal working language of Ethiopian, and other three are regional official languages. 
Amharic, Afan Oromo, Somali, and Tigrinya are the top four languages in Ethiopia by the number of speakers. 
~\\\textbf{Amharic} (amh): is a Semitic language written in Ge'ez script, known as Fidel, which consists of 33 primary characters, each with seven vowel sequences. It is the second most widely spoken Semitic language, next to Arabic. %Among the Amharic dialects, These are significant dialects: Gondar, Gojjami, and Showa. Specially marked differences exist in pronunciation, vocabulary, and grammar between the northern Gojjami and the southern Showa dialects.
~\\\textbf{Afan Oromo} (orm): is an Afro-Asiatic language written in Latin script. It is the most widely spoken language in Ethiopia and the third most widely spoken in Africa, next to the Arabic and Hausa languages. It is mostly spoken in the Horn of Africa, including Ethiopia, Kenya, and Somalia alone.
~\\\textbf{Somali} (som): is an Afro-Asiatic language belonging to the Cushitic group. It is spoken in Ethiopia, Somaliland, Kenya, and Somalia. It is the third most widely spoken language in Ethiopia.
~\\\textbf{Tigrinya} (tir): is a Semitic language spoken in the Tigray region of Ethiopia and Eritrea. The language uses Ge'ez script with additional Tigrinya alphabets and is closely related to Ge'ez and Amharic \cite{Eberhard2023}.

\section{Annotators Background }
\label{sec:appb}
\begin{figure*}[h]
    \centering
    \includegraphics[width=\linewidth]{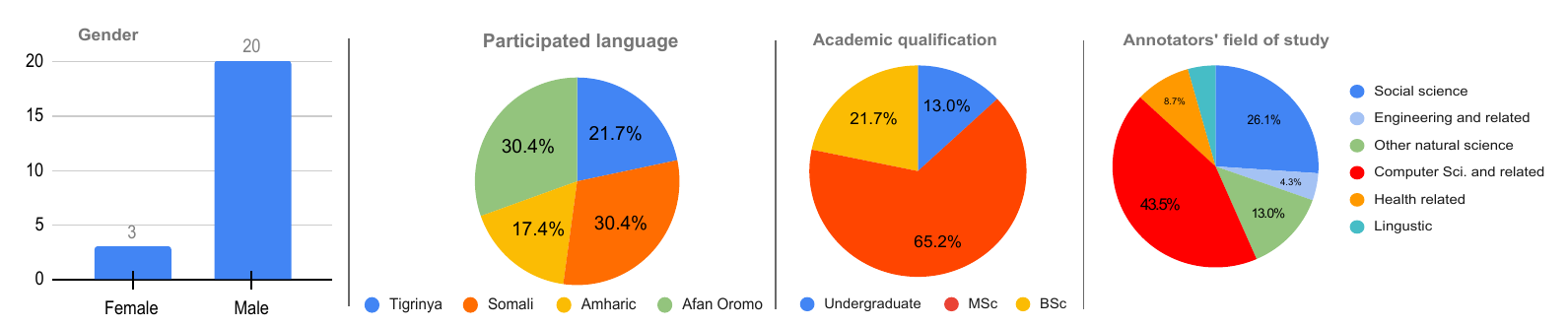}%=14cm, height=8cm
    \caption{Backgrounds of Annotators: gender, language participated, academic qualification, and field of study.}
    \label{fig:annot}
\end{figure*}
% \section{Annotation Tool Interface}
% \label{appendix:annotation_tool}
% Figure \ref{fig:annotool} shows the screenshot of the annotation interface. The tool includes the text to be annotated (one instance displayed at a time), emotion options with checkboxes to allow multiple selections of emotions except for the neutral class, forward and backward buttons, shortcuts to select emotions and annotator details. 
% \begin{figure*}[h]
%     \centering
%     \includesvg[width=\linewidth]{images/Screenshot_anno.svg}%=14cm, height=8cm
%     \caption{: The user interface screenshot with emotion annotation task comprises the emotion's source text and emotion choices}
%     \label{fig:annotool}
% \end{figure*}

% \clearpage
\section{Number of Emotion Labels Per Instance}
\begin{figure*}[h]
    \centering
    \includegraphics[width=\linewidth]{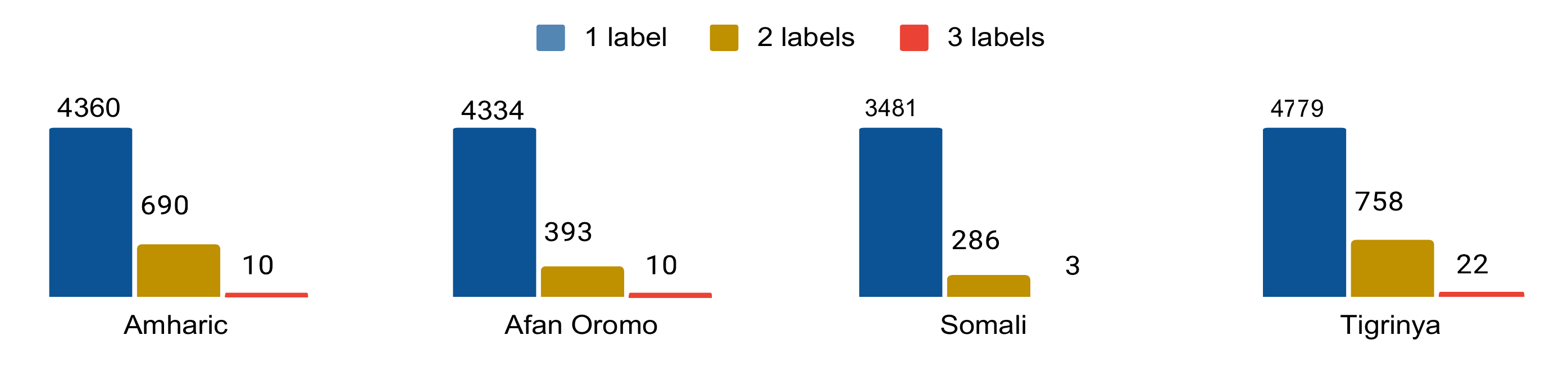}
    \caption{Number of emotion labels per instance across languages}
    \label{fig:label-dist}
\end{figure*}

Figure \ref{fig:label-dist} shows the number of emotion label(s) distribution across languages. As we can see, most of the dataset for all languages has a single emotion class. Of the \texttt{amh} dataset, 88\%  has a single label, 11.7\% has two emotion labels, and 0.17\% has three labels. In the \texttt{orm},  92.9\% has a single label, 6.9\% two labels, and  0.17\% three labels. In \texttt{som}, 94.8\% has single labels,  5.7\% two labels, and only three instances have three labels. In \texttt{tir}, 82.3\% has single labels, 12.4\% two labels, and 0.36\% three labels. \ethioemo dataset is also used for multiclass emotion classification for future work as instances with more than two labels are less.
\label{label-dist}

\clearpage
% \section{Prompts for Zero-shot and ICL}
% % \quad
% \label{appendex:zerofewshot}
% % Table \ref{tab:prompt_templates} shows examples of prompts used for decoder-only zero-shot and in-context learning experiments.
% \begin{table*}[h]
% % \vspace*{-10pt} 
% \centering
% % \begin{center}
% \scalebox{0.7}{
% \begin{tabular}{p{210mm}}
% \toprule
% \textbf{Prompts used for zero-shot and ICL evaluations} \\
% \midrule
% \textbf{Prompt 1}: \texttt{Identify all applicable emotions for the given text from (anger, disgust, fear, sadness, joy, surprise).} \\ 
% \texttt{Text: """\{text\}"""} \\                
% \texttt{Answer:}  \\ 
% \textbf{Prompt 2}: \texttt{Categorize the text's emotional tone as the presence of one or more of the given emotions (anger, disgust, fear, sadness, joy, surprise) only. No explanation is needed.} \\ 
% \texttt{Text: """\{text\}"""} \\                
% \texttt{Answer:}  \\ 
% \textbf{Prompt 3}: \texttt{You are a helpful AI assistant that can identify emotions from text. \newline
%                  Categorize the text’s emotional expression, classifying it as one or more of the specified emotions [\lq anger\rq, \lq disgust\rq, \lq fear\rq, \lq sadness\rq, \lq joy\rq, \lq surprise\rq] that reflect the writer’s state of mind. No explanation is needed.} \\ 
% \texttt{Text: """\{text\}"""} \\                
% \texttt{Answer:}  \\ 
% \addlinespace
% \bottomrule
% \end{tabular}
% }
% \caption{The three prompts used for decoder-only zero-shot and in-context learning experiments} 
% \label{tab:prompt_templates}
% % \end{center}
% \end{table*}

\section{Experiment Hyper-parameters}
We make fine-tune encoder-only models using FLAIR framework \cite{flair-2019} with the following hyper-parameters: model\_max\_length = 512 (except AfroLM model\_max\_length is 256 as the models built it up), learning\_rate = 5.0e-5, mini\_batch\_size = 8, and max\_epochs=3, as recommended in the BERT paper \cite{devlin2018bert}. We test decoder-only models with temperature = 0 and batch\_size = 1.

\section{Emotion Examples for Error Analysis}
\label{sec:app-error}
% Table \ref{fig:error} shows examples of emotional texts for each identified case discussed in the experiment error analysis section, Section\ref{sec:error_analysis}. 

\begin{figure*}[ht!]
    \centering
    \includegraphics[width=\linewidth]{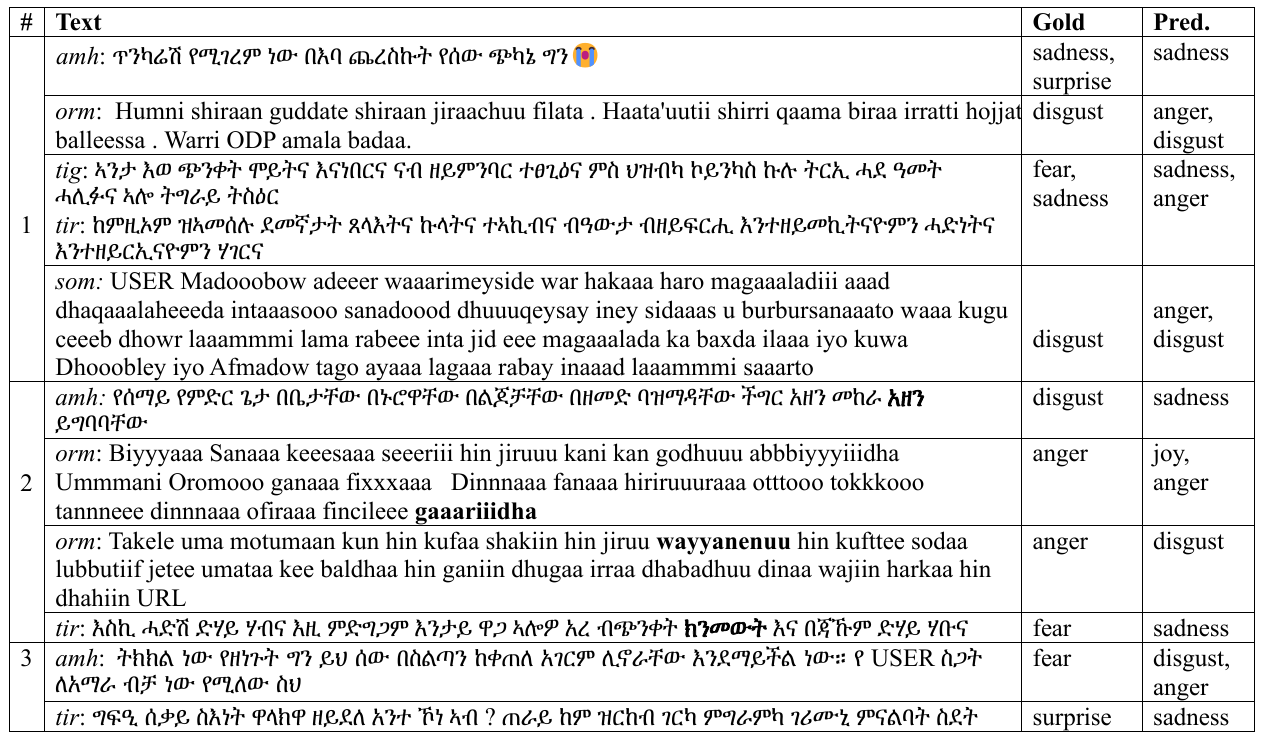}
    \captionof{table}{Examples of emotion texts for cases discussed in the experiment error analysis, Section \ref{sec:error_analysis}; \texttt{amh}, \texttt{orm}, \texttt{som}, and \texttt{tir} are the languages used in the example. }
    \label{fig:error}
\end{figure*}

% \clearpage
\section{Data Source and Emotion Distribution}
\label{sec:app-datasource}
\begin{figure}[h]
    \centering
    \includegraphics[width=\linewidth]{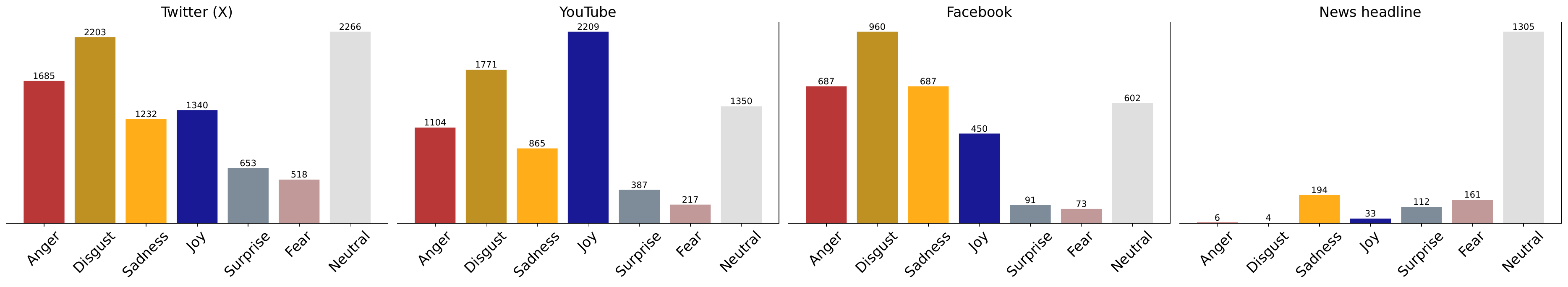}
    \caption{Emotion distribution in the data sources for \ethioemo dataset}
    \label{fig:dt}
\end{figure}

\begin{figure}[h]
    \centering
    \includegraphics[width=\linewidth]{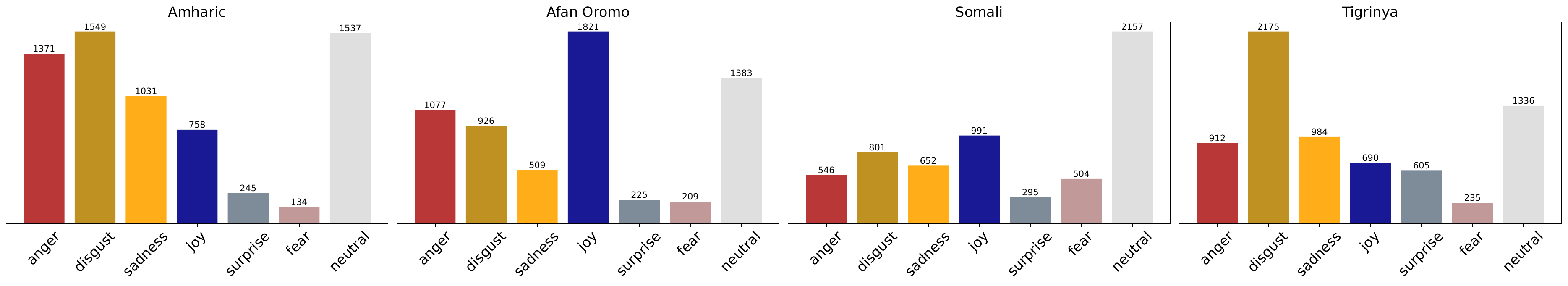}
    \caption{Emotion distribution for \ethioemo dataset across languages }
    \label{fig:emotion-distrb}
\end{figure}

\begin{figure}[h]
    \centering
    \includegraphics[width=\linewidth]{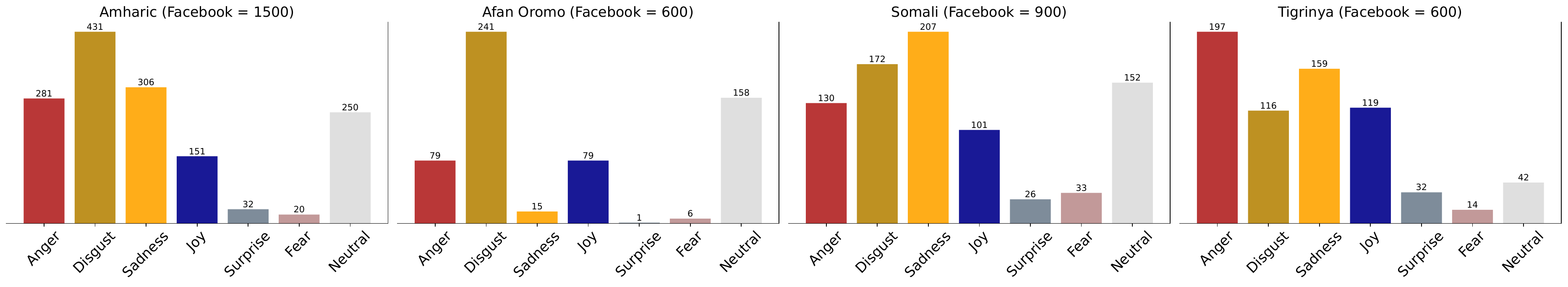}
    \caption{Facebook comment emotions distribution across languages from the given quota }
    \label{fig:f}
\end{figure}

\begin{figure}[hbt!]
    \centering
    \includegraphics[width=\linewidth]{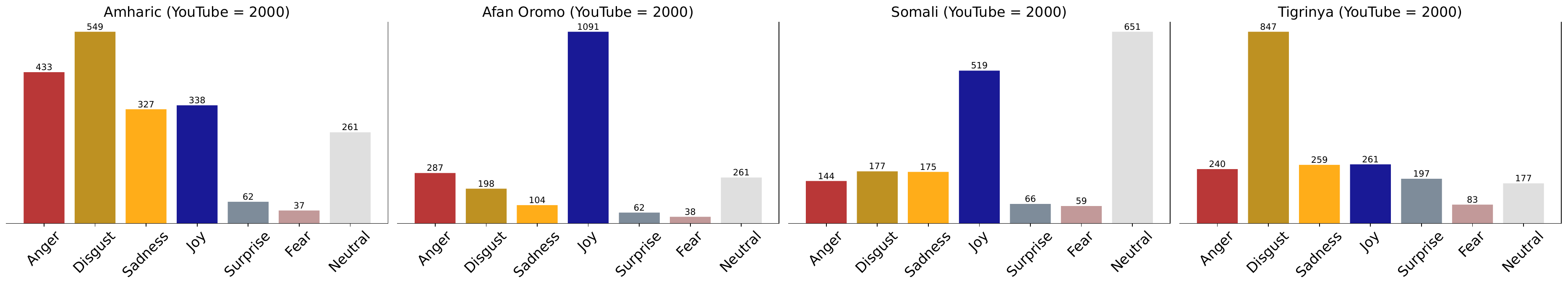}
    \caption{YouTube comment emotions distribution across languages from the given quota }
    \label{fig:y}
\end{figure}

\begin{figure}[hbt!]
    \centering
    \includegraphics[width=\linewidth]{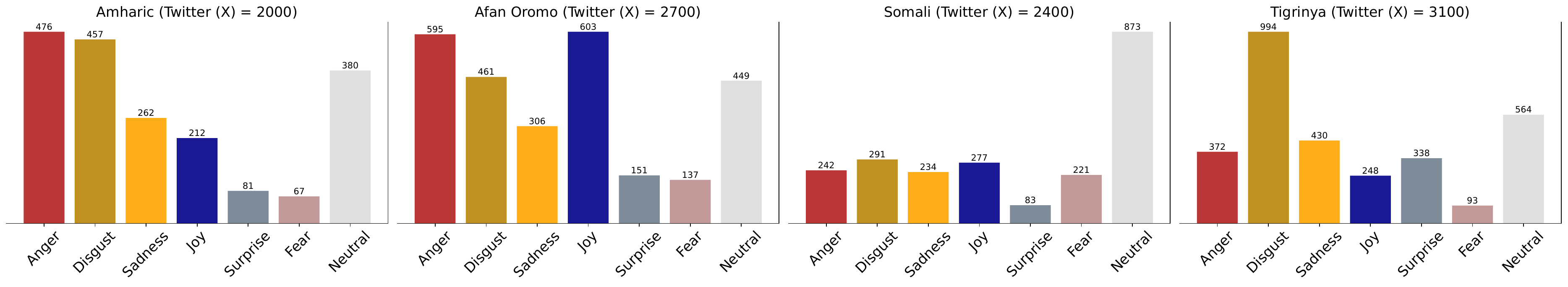}
    \caption{Twitter (X) post emotions distribution across languages from the given quota } 
    \label{fig:x}
\end{figure}

\begin{figure}[hbt!]
    \centering
    \includegraphics[width=\linewidth]{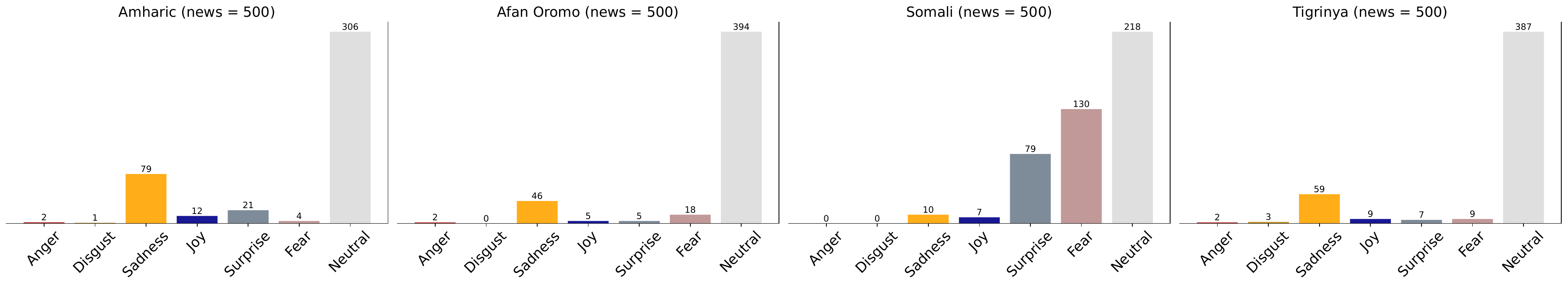}
    \caption{News headlines emotions distribution across languages from the given quota  }
    \label{fig:n}
\end{figure}

In this section, we visualize the \ethioemo dataset emotion distributions across data sources: Twitter (X), YouTube, Facebook, and news headlines and languages: \texttt{amh}, \texttt{orm}, \texttt{som}, and \texttt{tir}. Figure \ref{fig:dt} shows general emotion distribution across data sources for the \ethioemo dataset. Figure \ref{fig:emotion-distrb} shows emotion distribution across languages.  Figures \ref{fig:f}, \ref{fig:y}, \ref{fig:x}, and \ref{fig:n} show emotion distribution across languages and data sources: Facebook comments, YouTube comments, Twitter (X), and news headlines, respectively. Figure \ref{fig:n} shows that the news headlines almost do not have any of the basic emotions. %News headlines do not contain basic emotions in Ethiopian languages.

% \vspace*{-1in}
% \section{Emotion distribution across the data sources and languages}
    % \vspace*{3em}
    % \end{subfigure}
    % \vspace*{3em}
%     \vspace*{3em}
% \clearpage
%     \vspace*{3em}
%     \caption{Emotion distribution across the data sources and languages. Each pilot bar shows the statistics of each emotion after annotation from the corresponding data source. }
% \end{figure}

\clearpage

\section{Additional Results}
\subsection{Encoder-only Experimental Results}
\label{sec:appendix2}
We compared our experimental results using a weighted-averaged F1-score. Table \ref{tab:baseline2} shows additional multi-label evaluation metrics such as multi-label accuracy, macro F1-score, and micro F1-score across languages and encode-only models.

\begin{table*}[h]
% \footnotesize
\centering
\scriptsize
\begin{tabular}{lcccccccccccc}
\toprule
 & \multicolumn{3}{c}{\textbf{Amharic} (amh)} & \multicolumn{3}{c}{\textbf{Afan Oromo} (orm)}  & \multicolumn{3}{c}{\textbf{Somali} (som)} & \multicolumn{3}{c}{\textbf{Tigrinya} (tir)} \\
\cmidrule(rl){2-4}\cmidrule(rl){5-7}\cmidrule(rl){8-10}\cmidrule(rl){11-13}
 \textbf{Pre-trained LMs}  & \textbf{Acc} & \textbf{Mac F1} & \textbf{Mic F1}  & \textbf{Acc} &\textbf{Mac F1} & \textbf{Mic F1}  & \textbf{Acc} & \textbf{Mac F1} & \textbf{Mic F1}  & \textbf{Acc} &\textbf{Mac F1}&\textbf{Mic F1}  \\
\hline
EthioLLM-s-70K&50.3&58.8&69.0&60.94&61.4&70.0&40.8&42.4&51.0&48.0&46.3& 59.8\\
EthioLLM-l-70K&48.6&53.9&65.1&60.0&55.7&68.9&31.2&32.2&43.6&48.9&46.8& 60.3\\
% EthioLLM-l-250K&47.7&56.3&63.4&60.6&59.2&60.0&41.9&40.4&59.5&51.3&50.1& 60.4\\
Afro-xlmr-large-61L&54.0&68.3&68.4&58.4&55.8&68.0&53.4&61.7&64.9&51.4&55.4& 64.5\\
Afro-xlmr-large-76L&55.6&67.5&70.0&63.3&66.3&72.9&53.6&59.4&64.3&49.1& 48.4&61.4\\
AfroLM-active-learning&50.0&60.7&65.6&59.1&52.2&66.2&40.9&48.2&53.2&36.6&33.8& 49.5\\
Afriberta-large&67.5&64.2&67.7&62.4&63.7&71.8&53.7&60.1&64.5&49.3&52.8& 62.6\\
\bottomrule
\end{tabular}
\caption{Additional results of encoder-only models. Acc - multi-label average accuracy, Mac F1 - macro F1-score, and Mic F1 - micro F1-score.} 
\label{tab:baseline2}
\end{table*}

\subsection{Emotion Class-based Results}
\label{app:class-result}

Table \ref{tab:class} shows class-level emotion results from the three encoder-only models. We discovered that the emotion class with less dataset distribution performs low, for example, \texttt{Fear} class in \texttt{orm} and \texttt{tir} languages. In overall performance across languages, \texttt{som} and \texttt{tir} have low performance; this might be because of the amount of corpus in the pre-training and the data in the emotion classes.
\begin{table*}[!h]
% \footnotesize
\centering
\scriptsize
\begin{tabular}{lcccccccccccccccc}
\toprule
 & \multicolumn{4}{c}{\textbf{EthioLLM-s-70K}} & \multicolumn{4}{c}{\textbf{Afro-xlmr-large-61L}}  & \multicolumn{4}{c}{\textbf{Afro-xlmr-large-76L}} &\multicolumn{4}{c}{\textbf{AfriBERTa-large}}\\
\cmidrule(rl){2-5}\cmidrule(rl){6-9}\cmidrule(rl){10-13}\cmidrule(rl){14-17}
 \textbf{Emotions}  
 & \textbf{amh} & \textbf{orm} & \textbf{som}  & \textbf{tir} &\textbf{amh} & \textbf{orm} & \textbf{som}  & \textbf{tir}&\textbf{amh} & \textbf{orm} & \textbf{som}  & \textbf{tir}&\textbf{amh} & \textbf{orm} & \textbf{som}  & \textbf{tir} \\
 \hline
Anger   
&56.3&55.1&10.0&19.5 &59.6&55.1&41.1&25.6 &58.7&61.2&30.2&20.2 &58.3&57.4&39.7&29.2\\
Disgust 
&68.4&68.9&36.7&71.9 &66.7&629&58.3&77.2 &71.5&68.2&57.4&73.6 &68.6&68.1&53.8&74.6\\
Fear  
&21.7&30.2&57.8&2.8 &48.7&12.3&69.1&27.1 &53.9&42.0&70.7&00.0 &40.7&34.9&73.2&18.2\\
Sadness 
&74.7&55.2&53.7&53.9 &75.0&54.1&71.4&62.9 &77.8&63.1&68.7&58.4 &74.7&62.0&70.4&61.7\\
Joy  
&71.2&85.6&72.4&56.9 &82.3&84.2&77.5&63.2 &81.2&87.0&79.8&64.2 &76.9&87.5&79.0&60.9\\
Surprise
&60.6&73.0&23.9&73.0 &65.5&66.7&52.9&76.5 &62.3&76.0&49.6&74.1 &66.2&72.4&44.6&72.3\\
\bottomrule
\end{tabular}
\caption{Class-based emotion F1 results from selected fine-tuned encoder-only models}%\afriberta, \afroxlmrlargest, \ethiollmbase models} 
\label{tab:class}
\end{table*}

% \clearpage
% \section{Confusion matrix}
% \label{conf-matrix}
% Figure \ref{fig:cm-amh} and \ref{fig:cm-oro} show the confusion matrix from \textbf{Afro-xlmr-large-76L} model for Amharic and Afan Oromo languages, respectively. The confusion matrix in multi-label case shows the classification results of one emotion class versus the others (the remaining classes grouped as one class). Note that an instance in the multi-label dataset might have one or more classes, and the cells of the confusion matrix may have meanings different from those of the binary or multiclass confusion matrix. The first cell (box) in the matrix shows the correctly predicted statistics for the corresponding emotion class.
% \begin{figure*}[h]
%     \centering
%     \includegraphics[width=\linewidth]{images/cm_amh.png}
%     \caption{Amharic confusion matrix - OneVsRest from \textbf{Afro-xlmr-large-76L}. Test split emotion distributions are anger 408, disgust 461, fear 37, sadness 308, joy 226, surprise 70, and neutral 461.}
%     \label{fig:cm-amh}
% \end{figure*}

% \begin{figure*}[h]
%     \centering
%     \includegraphics[width=\linewidth]{images/cm_oro.png}
%     \caption{Afan Oromo Confusion matrix - OneVsRest from \textbf{Afriberta-large}. Test split emotion distributions are anger 318, disgust 272, fear 58, sadness 148, joy 542, surprise 66, and neutral 409.}
%     \label{fig:cm-oro}
% \end{figure*}

% \clearpage

\clearpage
\subsection{Results Across Prompts, Languages, and k-shots}
\label{sec:prompt}
The details of the prompts are shown in Figure \ref{fig:prompt}. Prompt 1 is a generic prompt, Prompt 2 is a task-based prompt, and Prompt 3 is a role-based prompt.
\begin{table*}[h!]
\centering
\scriptsize
\begin{tabular}{ccccccccccccccccc}
\toprule
 & &\multicolumn{5}{c}{\textbf{Gemma-2b-it}} & \multicolumn{5}{c}{\textbf{Gemma-1.1-7b}}  & \multicolumn{5}{c}{\textbf{LLaMA-2-7b-chat-hf}}\\
\cmidrule(rl){3-7}\cmidrule(rl){8-12}\cmidrule(rl){13-17}
  &\textbf{Lang} & \textbf{0} & \textbf{2} & \textbf{4}  & \textbf{6} &\textbf{8} & \textbf{0} & \textbf{2} & \textbf{4}  & \textbf{6} &\textbf{8}&\textbf{0} & \textbf{2} & \textbf{4}  & \textbf{6} &\textbf{8}  \\
\hline
% \parbox[t]{2mm}{\multirow{5}{*}{\rotatebox[origin=c]{90}{EthioLLM}}}
\multirow{5}{*}{\rotatebox[origin=c]{90}{\parbox[c]{1cm}{\centering Prompt 1}}} 
&amh&8.33&16.98&12.98&11.14&12.99 &31.55&20.88&17.11&19.03&20 &3.59&49.74&49.68&21.74&25.62\\
&eng&47.27&59.92&58.6&58.91&61 &67.5&64.45&58&57.51&58.72 &49.09&60.36&59.82&58.01&59.63\\
&orm&7.05&18.04&11.57&10.57&11.01 &31.92&21.4&19.79&24.21&28.39 &11.44&25.86&20.24&22.26&24.6\\
&som&14.16&16.12&14.88&15.47&14.81 &20.8&25.83&25.73&26.21&27.57 &13.92&28.13&25.26&29.14&29.38\\
&tir&7.82&18.48&16.98&15.59&15.52 &27.84&13.56&12.86&12.71&14.04 &2.5&16.42&16.17&20.21&20.91\\
\hline
\multirow{5}{*}{\rotatebox[origin=c]{90}{\parbox[c]{1cm}{\centering Prompt 2}}} 
&amh&11.57&10.21&9.78&8.2&8.36 &25.61&21.26&18.42&19.18&21.27 &23.83&18.87&19.52&21.17&21.53\\
&eng&18.82&48.21&47.69&52.49&57.36 &65&60.54&58.64&58.58&58.63 &54.16&61.52&61.49&60.56&62.98\\
&orm&6.99&6.64&7.77&7.55&8.17 &36.36&18.54&21.11&25.19&29.4 &23.99&25.33&21.23&23.08&27.47\\
&som&15.13&12.31&13.11&12.51&12.36 &28.73&25.07&25.57&26.61&27.41 &26.41&27.75&25.15&27.68&28.37\\
&tir&8.73&9.68&15.91&13.32&10.52 &15.6&12.55&12.93&13.42&14.79 &22.07&15.92&15.23&19.38&19.06\\
\hline
\multirow{5}{*}{\rotatebox[origin=c]{90}{\parbox[c]{1cm}{\centering Prompt 3}}} 
&amh&10.76&12.9&11.57&10.34&9.6 &26.65&22.23&18.88&20.8&22.8 &24.64&20.59&20.29&23.58&25.79\\
&eng&43.11&47.9&47.05&55.36&57.89 &66.49&63.74&59.98&59.14&60.31 &59.97&61.15&61.39&62.46&64.21\\
&orm&9.39&9.39&9.74&7.68&9.23 &36.32&18.76&22.89&25.07&28.95 &22.28&26.62&24.14&27.17&27.85\\
&som&13.5&16.32&16.94&15.45&16 &28.09&26.99&26.42&26.11&28.56 &25.82&25.57&24.63&25.82&27.12\\
&tir&7.06&12.22&12.55&10.75&10.22 &15.2&14.09&13.39&16.04&15.11 &14.34&17.29&18.67&20.2&20.64\\
\hline
\multicolumn{17}{c}{ } \\ 
 & &\multicolumn{5}{c}{\textbf{LLaMA-3-8B-Instruct}} & \multicolumn{5}{c}{\textbf{LLaMA-3.1-8B-Instruct}}  & \multicolumn{5}{c}{\textbf{CohereForAI\_\_aya-101}}\\
 \cmidrule(rl){3-7}\cmidrule(rl){8-12}\cmidrule(rl){13-17}
\midrule
\multirow{5}{*}{\rotatebox[origin=c]{90}{\parbox[c]{1cm}{\centering Prompt 1}}} 
&amh&15.38&29.22&27.92&31.51&32.79 &21.63&34.25&33.58&37.96&39.52 &49.98&54.7&55&56.59&57.54\\
&eng&64.82&66.22&64.82&63.54&64.41 &54.64&68.21&67.42&66.4&67.22 &63.88&65.97&67.42&67.93&67.16\\
&orm&25.9&33.43&33.77&37.65&39.12 &29.71&36.24&37.75&38.25&40.67 &31.71&48.8&53.47&53.73&54.75\\
&som&26.01&32.65&33.49&34.77&34.46 &23.19&34.99&38.55&37.63&40.31 &36.49&47.39&49.18&50.71&51.84\\
&tir&8.4&21.72&19.64&22.37&23.45 &8.4&20.52&19.25&21.41&22.64 &41.52&48.96&50.09&50.84&50.01\\
\hline
\multirow{5}{*}{\rotatebox[origin=c]{90}{\parbox[c]{1cm}{\centering Prompt 2}}} 
&amh&36.95&27.63&27.25&30.15&30.55 &32.73&33.9&34.24&35.65&37.78 &47.38&53.73&54.96&56.12&66.66\\
&eng&67.12&65.87&63.55&61.94&63.58 &65.46&66.47&66.49&66.57&67.39 &65.21&65.33&65.83&66.75&66.66\\
&orm&24.6&33.29&31.72&35.72&37.24 &31.83&37.24&37.9&39.36&41.25 &35.42&47.82&51.52&53.9&54.27\\
&som&30.22&31.63&31.34&34.55&34.89 &29.57&18.08&16.69&17.42&18.45 &45.32&49.35&49.06&51.66&51.33\\
&tir&31.32&17.72&18.88&19.86&22.39 &15.34&18.08&16.69&17.42&18.45 &38.58&47.86&49.04&50.67&49.37\\
\hline
\multirow{5}{*}{\rotatebox[origin=c]{90}{\parbox[c]{1cm}{\centering Prompt 3}}} 
&amh&32.22&27.95&24.9&28.44&28.83 &7.38&34.68&33.86&36.2&37.04 &49.03&53.29&53.7&53.7&50.58\\
&eng&68.27&66.57&64.66&63.45&63.85 &33.4&67.07&66.84&65.46&65.93 &69.52&69.11&70.26&69.71&70\\
&orm&30.24&35.06&34.71&35.52&36.94 &10.77&36.13&37.32&37.97&39.32 &33.82&48.15&52.4&52.21&49.8\\
&som&31.64&29.73&29.39&31.41&33.77 &13.44&34.45&36.81&37.38&38.38 &17.18&18.9&50.62&50.91&49.51\\
&tir&19.46&21.37&21.51&25.17&25.41 &7.11&22.59&20.72&22.69&20.98 &36.81&41.2&40.1&40.05&33.92\\

\midrule
\multicolumn{17}{l}{\textit{Test-set translated results }} \\ 
 & &\multicolumn{5}{c}{\textbf{Gemma-2b-it}} & \multicolumn{5}{c}{\textbf{Gemma-1.1-7b}}  & \multicolumn{5}{c}{\textbf{LLaMA-2-7b-chat-hf}}\\
 \cmidrule(rl){3-7}\cmidrule(rl){8-12}\cmidrule(rl){13-17}
 \midrule
\multirow{5}{*}{\rotatebox[origin=c]{90}{\parbox[c]{1cm}{\centering Prompt 1}}} 
&amh&33.29&36.1&35.57&37.83&40.6 &42.67&43.12&39.06&40.31&41.53 &24.7&43.7&41.27&61.65&46.01\\
&orm&29.7&34.7&40.41&41.74&44.54 &46.52&40.71&39.7&42.48&41.9 &22.64&43.79&41.16&44.45&44.71\\
&som&29.24&38.42&40.27&40.74&42.85 &42.71&47.29&45.73&45.26&46.01 &28.44&46.24&45.71&46.03&48.89\\
&tir&25.57&26.5&29.07&30.47&30.22 &33.43&32&31.6&31.65&33.11 &19.88&31.09&31.02&31.15&37.08\\

\midrule
\multirow{5}{*}{\rotatebox[origin=c]{90}{\parbox[c]{1cm}{\centering Prompt 2}}} 
&amh&33.77&31.67&32.71&35.64&37.45 &45.44&40.4&40.73&40.58&42.45 &41.92&43.39&41.75&43.43&45.51\\
&orm&28.87&31.1&34.75&36.01&40.65 &47.85&34.96&40.82&43.49&45.05 &38.36&43.07&44.4&45.88&45.71\\
&som&33.9&37.83&39.37&40.67&41.81 &46.32&45.39&45.89&45.64&47.73 &35.8&47.12&46.8&46.08&50.15\\
&tir&24.45&24.42&27.86&30.47&31.28 &37.6&30.69&32.15&33.11&34.82 &29.77&29.05&29.79&30.24&33.69\\

\midrule
\multirow{5}{*}{\rotatebox[origin=c]{90}{\parbox[c]{1cm}{\centering Prompt 3}}} 
&amh&25.68&34.1&34.09&35&38 &47.03&42.65&41.07&40.61&42.73 &39.82&41.93&40.81&42.54&45.17\\
&orm&27.42&33.17&35.77&37.66&40.61 &49.31&39.84&41.38&44.64&44.19 &41.82&41.81&41.2&44.75&46.43\\
&som&31.14&39.9&39.34&42.55&43.02 &45.12&46.25&46.1&46.51&47.22 &39.5&43.54&44.52&46.22&47.67\\
&tir&17.6&24.69&27.2&29.02&30.44 &35.02&30.85&31.06&33.2&34.02 &25.93&27.7&28.88&32.64&35.4\\

\midrule
\multicolumn{17}{c}{ } \\ 
 & &\multicolumn{5}{c}{\textbf{LLaMA-3-8B-Instruct}} & \multicolumn{5}{c}{\textbf{LLaMA-3.1-8B-Instruct}}  & \multicolumn{5}{c}{\textbf{CohereForAI\_\_aya-101}}\\
 \cmidrule(rl){3-7}\cmidrule(rl){8-12}\cmidrule(rl){13-17}
\midrule
\multirow{5}{*}{\rotatebox[origin=c]{90}{\parbox[c]{1cm}{\centering Prompt 1}}} 
&amh&45.77&47.4&45.87&45.44&46.06 &33.85&46.09&47.37&46.16&48.51 &42.27&47.81&49.62&49.86&49.27\\
&orm&47.65&47.96&46.3&46.83&47.38 &44.14&48.49&48.41&50.4&49.17 &39.86&44.72&45.55&46.66&47.6\\
&som&42.81&48&48.65&48.39&48.91 &36.12&47.04&47.9&49.73&50.66 &37.36&46.34&44.48&46.42&47.36\\
&tir&38&34.13&32.17&34.73&36.34 &27.84&33.58&37.05&39.74&40.63 &27.01&39.338&38.47&38.08&39.27\\

\midrule
\multirow{5}{*}{\rotatebox[origin=c]{90}{\parbox[c]{1cm}{\centering Prompt 2}}} 
&amh&48.12&45.55&44.69&45.31&45.13 &38.73&44.51&46.01&45.96&48.25 &47.38&48.62&48.72&49.51&47.85\\
&orm&46.94&47.27&45.64&46.36&47.99 &38.5&46.85&46.84&48.62&48.56 &43.76&45.71&45.5&47.21&48.03\\
&som&46.82&46.92&47.09&46.96&48.12 &48.03&45.76&47.74&49.69&49.51 &41.17&46.83&45.07&46.77&46.77\\
&tir&40.77&28.77&30.75&34.48&33.25 &28.11&30.47&35.8&36.47&38.28 &37.5&39.31&38.68&37.08&38.08\\
\midrule
\multirow{5}{*}{\rotatebox[origin=c]{90}{\parbox[c]{1cm}{\centering Prompt 3}}} 
&amh&50.89&46.61&44.85&45.04&46.18 &13.18&46.07&46.79&46.17&48.08 &45.2&46.79&46.54&44.39&44.71\\
&orm&51.05&45.72&44.78&45.46&46.52 &20.65&47.76&48.42&47.3&48.85 &46.56&47.93&49&48.19&47\\
&som&49.75&45.65&47.18&46.94&47.93 &20.7&46.89&48.23&48.9&48.48 &46.02&46.89&46.16&47.34&47.33\\
&tir&44.02&33.01&31.51&34.1&37.51 &9.02&34.28&37.26&38.66&40.05 &30.37&33&32.12&31.44&29.99\\
\bottomrule
\end{tabular}
\caption{Prompt sensitivity experiment results across k-shots, LLMs, and languages } 
\label{tab:prompts}
\end{table*}

\end{document}